\documentclass[sigconf]{acmart}

\AtBeginDocument{%
  }

\acmYear{2026}\copyrightyear{2026}
\setcopyright{cc}
\setcctype[4.0]{by}
\acmConference[MMSys '26]{ACM Multimedia Systems Conference 2026}{April 4--8, 2026}{Hong Kong, Hong Kong}
\acmBooktitle{ACM Multimedia Systems Conference 2026 (MMSys '26), April 4--8, 2026, Hong Kong, Hong Kong}
\acmDOI{10.1145/3793853.3799814}
\acmISBN{979-8-4007-2481-7/26/04}

\acmSubmissionID{80}

\usepackage{gensymb}
\usepackage[capitalise,noabbrev]{cleveref}
\usepackage{booktabs}
\usepackage{multirow}
\usepackage{colortbl}

\definecolor{pastelyellow}{RGB}{255,248,220}
\definecolor{pastelgreen}{RGB}{190,230,200}
\definecolor{mygreen}{RGB}{0,120,0}
\definecolor{mymyorange}{RGB}{200,100,0}

\usepackage[per-mode=symbol]{siunitx}
\usepackage{xspace}
\usepackage[table,xcdraw]{xcolor}
\usepackage{enumitem}
\usepackage[normalem]{ulem}
\usepackage{placeins}
\setlist[itemize]{leftmargin=*}

\makeatletter
\DeclareRobustCommand\onedot{\futurelet\@let@token\@onedot}
\def\@onedot{\ifx\@let@token.\else.\null\fi\xspace}
\DeclareRobustCommand\nodot{\futurelet\@let@token\@nodot}
\def\@nodot{\ifx\@let@token.\else~\null\fi\xspace}

\newfont{\eaddfnt}{phvr8t at 12pt}
\def\email#1{{{\eaddfnt{\par #1}}}}

\def\eg{\emph{e.g}\onedot} 
\def\vs{vs\onedot} 

\def\ie{\emph{i.e}\onedot} 
 
 \def\vs{\emph{vs}\onedot}

\def\etal{\emph{et al}\onedot}

\DeclareSIUnit\px{px}
\newcommand{\commentaire}[1]{\textcolor{#1}}

\newcommand{\gmr}{\commentaire{purple}} 

\newcommand{\brand}[1]{\textsf{#1}\xspace}

\newcommand{\name}{\brand{MoonAnything}}
\newcommand{\lunarstereo}{\brand{LunarGeo}}
\newcommand{\lunargr}{\brand{LunarPhoto}}
\newcommand{\lunarphoto}{\lunargr}

\begin{document}

\title{{\name}: A Vision Benchmark\\ with Large-Scale Lunar Supervised Data}
\renewcommand{\shorttitle}{{\name}: A Vision Benchmark with Large-Scale Lunar Supervised Data}




\author{Clémentine Grethen$^{1}$, Yuang Shi$^{1,2,3}$, Simone Gasparini$^{1,3}$, Géraldine Morin$^{1,3}$}
\def \authors{Clémentine Grethen et al.}
\affiliation{%
\institution{$^1$IRIT - Université de Toulouse \country{France} \quad $^2$National University of Singapore \country{Singapore} \quad $^3$IPAL, IRL2955 \country{Singapore}}
}

\renewcommand{\shortauthors}{Grethen \etal.}


\begin{abstract}

Accurate perception of lunar surfaces is critical for modern lunar exploration missions.
However, developing robust learning-based perception systems is hindered by the lack of datasets that provide both geometric and photometric supervision.
Existing lunar datasets 
typically lack either geometric ground truth, photometric realism, illumination diversity, or large-scale coverage.
In this paper, we introduce {\name}, a unified benchmark built on real lunar topography with physically-based rendering, providing the first comprehensive geometric and photometric supervision under diverse illumination with large scale.
The benchmark comprises two complementary sub-datasets 
: i) {\lunarstereo} provides stereo images with corresponding dense depth maps and camera calibration enabling 3D reconstruction and pose estimation;
ii) {\lunargr} provides photorealistic images using a spatially-varying BRDF model, along with multi-illumination renderings under real solar configurations, enabling reflectance estimation and illumination-robust perception.
Together, these datasets offer over 130K samples with comprehensive supervision.
Beyond lunar applications, {\name} offers a unique setting and challenging testbed for algorithms under low-textured, high-contrast conditions and applies to other airless celestial bodies and could generalize beyond.
We establish baselines using state-of-the-art methods and release the complete dataset along with generation tools
to support community extension: \url{https://github.com/clementinegrethen/MoonAnything}.

\end{abstract}




\maketitle


\begin{table*}[hbpt]
    \centering
    \caption{Comparison of lunar perception datasets. {\name} is the first to provide comprehensive geometric and photometric supervision with diverse illumination.}
    \vspace{-1em}
    \label{tab:dataset_comparison}
    \resizebox{\textwidth}{!} {%
        \begin{tabular}{llccccccc}
            \toprule
            \textbf{Dataset} & \textbf{Type} & \textbf{Region} & \textbf{Stereo} & \textbf{Depth} & \textbf{Reflect.} & \textbf{Multi-Light} & \textbf{Phys. Render} & \textbf{Scale} \\
            \midrule
            LRO NAC~\cite{vondrak2010lunar} & Real/Orbital & Global & \texttimes & \texttimes & \texttimes & \texttimes & N/A & 2M images \\
            Chang'E~\cite{wang2024change} & Real/Surface & Global & \texttimes & \texttimes & \texttimes & \texttimes & N/A & 7.5K images \\
            DLR’s TRON~\cite{lebreton2024trainingdatasetsgenerationmachine} & Lab & Mockup & \checkmark & \texttimes & \texttimes & \texttimes & \texttimes & 7.2K images \\
            POLAR~\cite{wong2017polar} & Lab & Mockup & \checkmark & \checkmark & \texttimes & \texttimes & \texttimes &  2.6K pairs \\
            LuSNAR~\cite{liu2024lusnar} & Synthetic & Simulated & \checkmark & \checkmark & \texttimes & \texttimes & \texttimes &  13K images \\
            StereoLunar~\cite{grethen2025adapting} & Synthetic 
            & South Pole & \checkmark & \checkmark & \texttimes & \texttimes & \checkmark &  50K pairs \\
            LunarG2R~\cite{grethen2026lunar} & Real & Tycho Crater& \texttimes & \texttimes & \checkmark & \texttimes & \checkmark &  83K pairs + DEM 
            \\
            \midrule
            {\name} & Real\&Synth. & Tycho\&S. Pole & \checkmark & \checkmark & \checkmark & \checkmark & \checkmark & 130K pairs + DEMs + SVBRDF
            \\
            \bottomrule
        \end{tabular}
    }%
\end{table*}

\section{Introduction}

Accurate perception of lunar surfaces is fundamental to modern lunar exploration.
With renewed global interest from NASA's Artemis program, China's Chang'E missions, and commercial ventures~\cite{niles2025commercial,wang2024change,goswami2009chandrayaan}, three perception capabilities become critical: \textit{Terrain Relative Navigation} (TRN) for precise landing without GPS, \textit{Hazard Detection and Avoidance} (HDA) for identifying unsafe terrain during descent, and \textit{3D surface reconstruction} for rover navigation~\cite{johnson2008overview,johnson2008analysis,herbort2011introduction}.
These capabilities rely on computer vision algorithms that operate reliably under the Moon's challenging conditions.

The lunar environment presents challenges for visual perception. 
The lack of atmosphere creates pitch-black shadows with extreme contrast~\cite{bickel2021peering}, while the textureless regolith surface causes traditional feature detectors to fail~\cite{posada2024dense,grethen2025adapting}.
These factors create a substantial domain gap, thus deep learning models trained on Earth imagery fail on lunar scenes, producing unreliable geometry estimations with flat reconstructions and inconsistent relief~\cite{grethen2025adapting}.

High-quality datasets are essential for developing robust lunar perception systems. 
Recent 3D reconstruction models such as DUSt3R~\cite{wang2024dust3r}, MASt3R~\cite{leroy2024mast3r}, and VGGT~\cite{wang2025vggt} achieve remarkable performance on human-scale imagery precisely because they are trained on millions of real-world and synthetic images spanning indoor, outdoor, and object-centric scenes.
However, these models do not generalize to lunar imagery without domain-specific adaptation.
Although fine-tuning on lunar data has proven effective~\cite{grethen2026lunar}, such adaptation requires comprehensive datasets that capture the unique characteristics of lunar perception. 
%
%
Effective lunar vision datasets must provide three interrelated information: 
\begin{itemize}
    \item \textit{Geometry supervision} through stereo pairs and dense ground truth depth maps enables training of 3D reconstruction models that are then able to infer lunar terrain structure. 
    \item \textit{Appearance modeling} through an accurate surface reflectance model is essential for generating photorealistic simulations that support algorithm development and validation. 
    \item \textit{Illumination variation} is critical because lunar missions must operate across diverse lighting conditions, from the harsh shadows of polar regions to varying sun angles throughout the lunar day. Datasets that capture only single lighting conditions produce models that fail when deployed under different illumination.
\end{itemize}

Existing lunar datasets address these requirements only partially. 
Real mission data from NASA's Lunar Reconnaissance Orbiter (LRO) and China's Chang'E program provide valuable orbital and surface imagery, but lack the dense geometric ground truth required for supervised learning~\cite{vondrak2010lunar,wang2024change}. 
Synthetic datasets offer controlled supervision but have significant limitations. 
Generating photorealistic lunar imagery requires accurate modeling of surface reflectance through a Bidirectional Reflectance Distribution Function (BRDF), which describes how light reflects from a surface as a function of illumination and viewing angles.
For the Moon, this is particularly important because lunar regolith exhibits unique optical properties, including opposition surge and anisotropic scattering~\cite{sato2014hapke}, that differ substantially from terrestrial materials.
Furthermore, real lunar surfaces are not homogeneous: reflectance varies spatially across the terrain, requiring Spatially-Varying BRDF (SVBRDF) models for accurate rendering.
For example, LuSNAR~\cite{liu2024lusnar} provides stereo pairs with depth maps and semantic labels using Unreal Engine, but lacks physically accurate BRDF, limiting photometric realism.
%
The Polar Optical Lunar Analog Reconstruction (POLAR) stereo dataset~\cite{wong2017polar} offers laboratory-controlled imagery with precise ground truth, but covers only a limited size terrain mockup under fixed lighting. 

Our prior works partially addressed these challenges through two datasets: StereoLunar~\cite{grethen2025adapting} provides stereo imagery with depth supervision for the South Pole, while LunarG2R~\cite{grethen2026lunar} provides geometry and reflectance pairs for Tycho crater.
However, these datasets remain isolated. StereoLunar lacks reflectance supervision, LunarG2R lacks depth ground truth, and neither provides multi-illumination variation.
Critically, \textit{no existing dataset provides comprehensive geometric and photometric supervision under systematic illumination variation}.

In this paper, we introduce {\name}, a comprehensive benchmark for lunar surface perception that unifies and extends our previous works to address the limitations discussed above. {\name} comprises two complementary sub-datasets:
\begin{itemize}
    \item {\lunarstereo} is designed for geometric perception, providing stereo imagery with dense depth supervision across diverse viewing conditions and lunar regions.
    It extends LunarStereo~\cite{grethen2025adapting} by adding stereo imagery of the Tycho crater region, complementing the original South Pole coverage and enabling cross-region generalization studies.
    
    \item {\lunargr} is designed for appearance modeling, providing learned reflectance parameters and multi-illumination renderings that capture spatially-varying surface properties. 
    It extends LunarG2R~\cite{grethen2026lunar} with multi-illumination renderings generated using learned SVBRDF parameters, supporting photometric stereo and illumination-robust training.
\end{itemize}

{\name} provides the first unified benchmark offering large-scale geometric and photometric supervision under diverse illumination (see \cref{tab:dataset_comparison}). 
The dataset comprises over 130K pairs with corresponding DEMs and SVBRDF models (\textit{an order of magnitude larger than existing lunar datasets}), establishing a new standard for data-driven lunar perception research.
Beyond lunar tasks, {\name} serves broader computer vision research. The SVBRDF data enables material appearance modeling, while the textureless, high-contrast imagery provides a challenging benchmark for evaluating algorithm robustness under edge-case conditions applicable to other airless celestial bodies and low-feature environments.
Experiments establish baselines using state-of-the-art methods and demonstrate the dataset's utility for advancing lunar perception. 


The rest of the paper is organized as follows.
\Cref{sec:related} reviews existing lunar datasets and their limitations.
\Cref{sec:data_gen} details the data generation pipeline.
\Cref{sec:dataset} presents the dataset statistics and structure.
Finally, we demonstrate baseline experiments using SOTA methods in \Cref{sec:3d_reconstruct} and conclude in \Cref{sec:conclusion}.

\section{Related Work} \label{sec:related}

Lunar perception datasets fall into three main categories: i) synthetic simulations, ii) real mission imagery, and iii) laboratory mockups. 
We review each category and identify the gaps in the following.

\textbf{Synthetic simulation datasets} use graphics engines or planetary renderers to generate labeled lunar scenes with controlled ground truth. The Artificial Lunar Landscape dataset~\cite{pessia2019artificial} employs Terragen to create approximately \num{10000} photorealistic images with pixel-wise semantic labels for sky, small rocks, and large rocks. 
However, it provides no geometric ground truth, such as depth maps or camera poses. 
LuSNAR~\cite{liu2024lusnar} uses Unreal Engine to generate high-resolution stereo pairs along with depth maps, semantic labels, and simulated LiDAR/IMU data, offering a more comprehensive sensor suite. 
Space-specific tools such as PANGU~\cite{parkes2004pangu} and  SurRender~\cite{lebreton2024high} provide physically-based rendering capabilities tailored to planetary environments.
Despite their utility, these synthetic datasets 
typically lack real lunar topography, relying instead on procedurally generated or artist-designed terrain. 
Viewpoint diversity is often restricted to ground-level or near-nadir trajectories. 
Most critically, simplified lighting and reflectance models fail to capture the complex photometric behavior of lunar regolith. 

\textbf{Real lunar image datasets} provide authentic lunar appearance but 
lack geometric supervision.
The Chang'E dataset~\cite{wang2024change} aggregates $\approx$~\num{7500} descent-phase images from the Chang'E-3 and Chang'E-4 landers.
While these images provide an authentic representation of the lunar surface, they lack accurate 3D ground truth and calibrated camera poses.
Chandrayaan-2's Orbiter High Resolution Camera acquired stereo triplet sequences at $\approx$~\SI{0.3}{\meter\per\px} resolution~\cite{ISRO_Chandrayaan2}, yet image calibration parameters, precise poses, and aligned terrain models remain unpublished. 
NASA's Lunar Reconnaissance Orbiter (LRO) Narrow Angle Camera (NAC) delivers very high-resolution panchromatic imagery down to \SI{0.5}{\meter\per\px} ~\cite{robinson2011lro}, but these narrow swaths have sparse overlap and limited stereo baselines.
A fundamental limitation of real lunar datasets is 
that 
orbital and descent imagery is predominantly nadir-viewing with minimal lateral baseline, 
making these datasets unsuitable for supervised 3D perception because of the lack of 3D ground truth.

\textbf{Laboratory mockup datasets} 
provide controlled imagery with precise ground truth over physical terrain mockups.
The German Aerospace Center (DLR)'s TRON facility captures images over a \qtyproduct[product-units = single]{4x2}{\metre} lunar surface mockup using a robotic arm with millimeter-accurate positioning~\cite{lebreton2024trainingdatasetsgenerationmachine}.
NASA's POLAR dataset~\cite{wong2017polar} provides  $\approx$~\num{2500} HDR stereo pairs captured over a regolith simulant scene under low-angle illumination mimicking polar lighting conditions. 
The Synthetic Lunar Terrain (SLT) dataset~\cite{marcus2024slt} introduces neuromorphic event camera data alongside conventional imagery, expanding the sensor modalities available for lunar perception research.
While laboratory datasets offer precise geometric ground truth and repeatable conditions, they suffer from limited scale and diversity, with fixed or limited lighting conditions that do not represent the diversity of real lunar landscapes. 

\textbf{Summary and Positioning.} 
\cref{tab:dataset_comparison} summarizes the existing datasets. 
Synthetic datasets provide scalable labeled data but lack physical realism. 
Real datasets offer an authentic appearance but no geometric supervision. 
Laboratory datasets deliver precise ground truth but limited diversity. 
None provides comprehensive geometric and photometric supervision under illumination variation.
%
{\name} addresses this gap by using the real lunar Digital Elevation Models (DEMs).
Our proposed dataset provides two different settings: i) {\lunarstereo} provides stereo imagery with dense depth maps rendered using physically-based ray tracing over real lunar topography,
spanning diverse trajectories over two geographic regions; and ii) \lunargr provides images rendered with SVBRDF parameters learned from real satellite imagery, along with multi-illumination renderings for material appearance modeling research.

Together, {\name} offers geometric realism grounded in actual lunar terrain, dense supervision for both geometry and illumination, and diverse lighting conditions. 
This comprehensive coverage enables training and evaluation of learning-based approaches for vision-based navigation, including terrain-relative navigation, hazard detection and avoidance, and 3D surface reconstruction.
\begin{figure*}[t!]
    \centering
    \includegraphics[width=1.\linewidth]{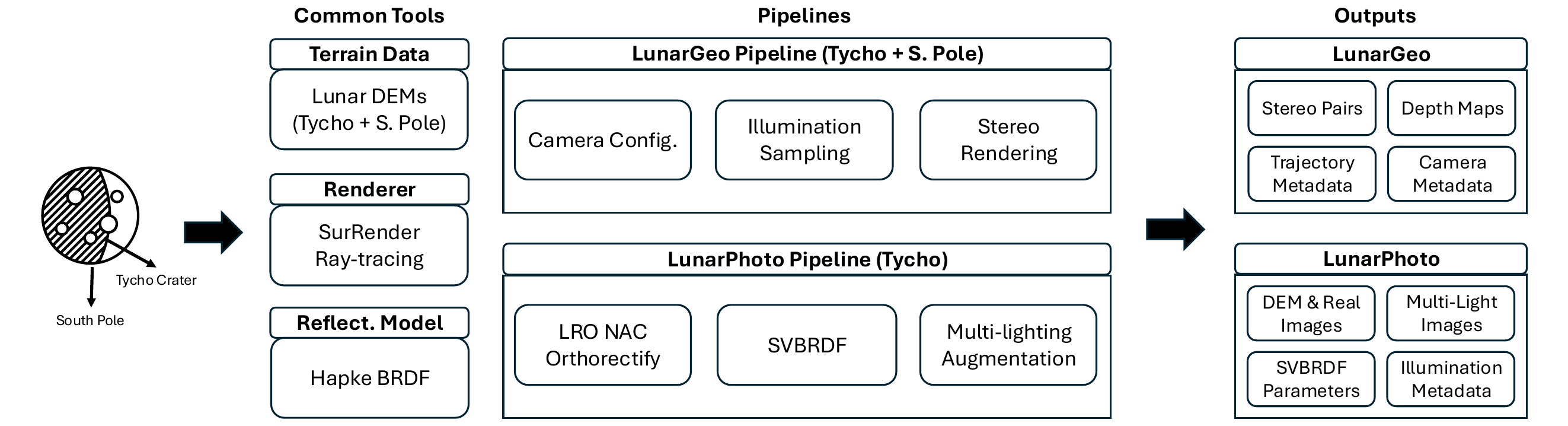}
    \caption{Overview of the {\name} dataset generation pipeline.}
    \label{fig:lunar_pipeline}
\end{figure*}

\section{Data Generation} \label{sec:data_gen}

\subsection{Common Pipeline Tools}

Both {\lunarstereo} and {\lunargr} share a common foundation of terrain data, rendering infrastructure, and reflectance modeling, as shown in \cref{fig:lunar_pipeline}. 
We first describe these shared components before detailing the specific generation procedures for each sub-dataset.

\textbf{Geographic Coverage.} 
{\name} spans two lunar regions: the \textit{South Pole} and the \textit{Tycho crater}. 
The South Pole is the target for Artemis III and presents the most challenging lighting conditions for lunar perception, with the Sun never rising more than \ang{7} above the horizon.
The Tycho crater, a well-studied mid-latitude impact site, offers dense multi-angle LRO imagery required for BRDF estimation. 
This geographic split mirrors real mission constraints, where perception methods must generalize across regions.

\textbf{Terrain Data Sources.} Both sub-datasets derive geometry from high-resolution Digital Elevation Models (DEMs):
\begin{itemize}
    \item \textit{South Pole}: A DEM at \SI{5}{\meter\per\px} resolution derived from LOLA laser altimeter data captured by NASA's LRO mission \cite{Barker2016}. This region features polar craters with elevations ranging from \SI{-4350}{\meter} to \SI{+1850}{\meter} relative to the lunar mean radius, including permanently shadowed basins and illuminated ridges.
    \item \textit{Tycho Crater}: A DEM at \SI{1}{\meter\per\px} resolution covering the Tycho crater region, produced using Airbus Pixel Factory from LRO stereo imagery~\cite{robinson2011lro}.
    This young impact crater exhibits pronounced geomorphological features, including a prominent central peak, terraced inner walls, and extensive ejecta deposits.
    The covered area spans approximately \SI{95}{\kilo\meter} \(\times\) \SI{90}{\kilo\meter}, with elevations ranging from \SI{-3570}{\meter} to \SI{1856}{\meter}.
    
\end{itemize}

\textbf{Rendering Framework.}
As illustrated in \cref{fig:lunar_pipeline}, both sub-datasets use a physically-based rendering pipeline built on the SurRender~\cite{lebreton2024high}. 
This framework integrates terrain geometry, surface reflectance, solar illumination, and camera models to produce imagery with full geometric supervision and calibrated metadata.

\textbf{Base Reflectance Model.}
Surface reflectance is modeled using the Hapke BRDF~\cite{hapke1993theory}, a physically-grounded formulation for airless bodies that captures opposition surge and anisotropic scattering. For the Tycho crater, we extend the standard constant-parameter Hapke model with a more elaborate SVBRDF model~\cite{grethen2026lunar} for improved photorealism.
{\lunarstereo} uses constant Hapke parameters for the South Pole and learned SVBRDF for Tycho; {\lunargr} uses the SVBRDF both for multi-lighting rendering and as ground truth for reflectance estimation.

\subsection{{\lunarstereo} for Geometric Perception} 
\label{subsec:lunarstereo}
{\lunarstereo} provides stereo image pairs with dense depth supervision for 3D reconstruction research and training. 
It is generated using real Lunar data that covers both the South Pole and Tycho crater regions, using the common rendering pipeline with region-specific camera, illumination, and reflectance configurations. 
For the South Pole, we render using the Hapke BRDF with constant albedo, providing consistent photometric behavior across the region.
For Tycho crater, we employ both the base Hapke model and the SVBRDF by using LunarG2R framework ~\cite{grethen2026lunar}.

\textbf{Camera Configuration}. 
Each stereo pair is rendered at \qtyproduct[product-units = single]{512 x 512}{\px}, with optical blur simulated via Gaussian point-spread function sampling. 
We use \ang{45} FoV for South Pole and \ang{30} FoV for Tycho (narrower to mitigate DEM artifacts).
We simulate three trajectory types inspired by lunar descent phases, informed by analysis of Chang'E-3 mission data and other descent studies~\cite{YU2014,Wong2006,Getchius2024}:

\begin{itemize}
    \item \textbf{Nadir:} Vertically downward-looking cameras simulating controlled descent. Stereo baselines range from {\SIrange{2}{10}{\percent}} of altitude, with both cameras at equal height.
    
    \item \textbf{Oblique:} Tilted cameras with viewing angles between {\ang{20} and \ang{35}}, simulating lateral motion or target-centered reorientation. Camera altitudes may be equal or offset.
    
\item \textbf{Dynamic:} Challenging configurations with altitude variation up to \SI{\pm 15}{\percent}, roll angles up to \ang{\pm 10}, and stereo baselines ranging from \SIrange{2}{22}{\percent} of the camera altitude.

\end{itemize}

\begin{table}[t]
    \centering
    \caption{Altitude versus effective GSD for {\lunarstereo}.}
    \vspace{-1em}
    \label{tab:gsd_vs_alt}
    \resizebox{\columnwidth}{!} {%
        \begin{tabular}{lcccccccccc}
            \toprule
            \multicolumn{11}{c}{\textbf{South Pole}}\\
            \midrule
            \textbf{Alt.} (\si{\kilo\meter}) & 3.5 & 6.2 & 9.5 & 12.8 & 16.1 & 19.4 & 22.7 & 26.0 & 29.2 & 30.5 \\
            \textbf{GSD} (\si{\meter\per\px}) & 5.7 & 10.0 & 15.4 & 20.7 & 26.0 & 31.4 & 36.7 & 42.1 & 47.2 & 49.3 \\
            \midrule
            \multicolumn{11}{c}{\textbf{Tycho Crater}}\\
            \midrule
            \textbf{Alt.} (\si{\kilo\meter}) & 3.5 & 4.1 & 4.7 & 5.3 & 5.9 & 6.5 & 7.1 & 7.7 & 8.3 & 8.9 \\
            \textbf{GSD} (\si{\meter\per\px}) & 3.7 & 4.3 & 4.9 & 5.5 & 6.2 & 6.8 & 7.4 & 8.1 & 8.7 & 9.3 \\
            \bottomrule
            
        \end{tabular}
    }%
\end{table}

Stereo pairs span region-specific altitude bands, yielding a range of ground sampling distances (GSDs), as summarized in \cref{tab:gsd_vs_alt}.
For the South Pole region, camera altitudes are sampled from \SI{3.5}{\kilo\meter} to \SI{30.5}{\kilo\meter}, covering a broad set of observation scales.
For the Tycho crater region, we instead consider 10 discrete altitude bands ranging from \SI{3.5}{\kilo\meter} to \SI{9.5}{\kilo\meter}.
This reduced altitude range limits the spatial footprint of each rendered image, preventing excessively large ground coverage and thereby mitigating the influence of DEM artefacts and interpolation errors on the rendered stereo pairs.
Camera positions are uniformly sampled across each region's spatial extent to ensure terrain diversity.

\textbf{Illumination Variation.}
To enable the study of illumination effects on reconstruction quality, each scene is rendered under three distinct realistic solar configurations to produce different shadow patterns. For the South Pole, we simulate the low-angle conditions typical of polar regions. For Tycho crater, we sample a broader range of illumination angles, reflecting its mid-latitude location.

\textbf{Ground Truth and Annotations.}
Each stereo pair includes comprehensive geometric and calibration annotation: 
i) \textit{Camera intrinsics:} Focal length, principal point, and sensor dimensions;
ii) \textit{Camera extrinsics:} 6-DOF poses in Moon-fixed reference frame;
iii) \textit{Dense depth maps:} Per-pixel depth along camera rays;
iv) \textit{Stereo geometry:} Inter-camera baseline translation and rotation;
and v) \textit{Trajectory metadata:} Altitude, GSD, and geographic coordinates.


\textbf{Geometric Perception Tasks.} 
The stereo pairs with dense depth supervision support stereo matching, multi-view 3D reconstruction, and camera pose estimation.
The diverse viewing configurations (nadir, oblique, dynamic) and altitude ranges make {\lunarstereo} suitable for evaluating algorithms under realistic descent conditions.
These capabilities directly support mission-critical applications, such as \emph{hazard detection and avoidance}, where accurate depth estimation enables the identification of rocks, craters, and unsafe slopes, and \emph{terrain-relative navigation}.

\subsection{{\lunargr} for Photometric Perception} \label{subsec:lunarg2r}

{\lunargr} provides paired geometry and reflectance 
data for appearance modeling research. 
It extends the original LunarG2R dataset~\cite{grethen2026lunar}, which provided geometry and reflectance pairs but only single-illumination observations per sample.
To the best of our knowledge, this is the first dataset providing spatially-varying BRDF parameters with diverse multi-illumination supervision.
The sub-dataset covers the Tycho crater region, combining real LRO observations with physically-grounded multi-lighting augmentation.

\textbf{Real Observation Extraction.}
Each sample consists of a DEM crop of \qtyproduct[product-units = single]{128 x 128}{\px} at \SI{5}{\meter\per\px} (\SI{0.4}{\kilo\meter\squared}) from the Tycho DEM (the same terrain source and resolution used in {\lunarstereo}).
For each crop, a corresponding LRO NAC~\cite{robinson2011lro} image (\SIrange{0.5}{2}{\meter\per\px}) is orthorectified onto the DEM and cropped to exactly match the spatial extent, producing a ground-truth appearance image aligned with the topography.
Metadata, including camera pose, Sun illumination direction, and geographic footprint, is given for each pair.

To construct valid training samples, we compute validity masks for both DEM and orthoimage sources, then erode the merged mask using a kernel equal to half the crop footprint to ensure complete spatial coverage. 
This defines the set of \emph{valid positions}, \ie, pixels that can safely serve as centers of crops whose full spatial extent contains only valid DEM and appearance data. 
Crop centers are randomly sampled from valid positions within a validity mask, with the number of crops per orthoimage proportional to its valid surface area.
Since orthoimages exhibit non-negligible spatial overlap, we perform geographic splitting, where the global DEM is divided into tiles, and all pairs whose centers fall within the same tile are assigned to the same train/val/test split, to avoid data leakage.

\textbf{SVBRDF.}
While the Hapke model provides physically-grounded reflectance, real lunar surfaces exhibit spatially-varying photometric properties not captured by constant-parameter assumptions.
We employ a learned SVBRDF model that captures opposition surge and backscattering effects, thus improving photorealism~\cite{grethen2026lunar}.

\textbf{Multi-Lighting Augmentation.}
Real LRO observations capture each terrain location under only limited illumination conditions.
To support tasks requiring diverse lighting (\eg, photometric stereo, intrinsic decomposition, and relighting), we augment the dataset with appearance images rendered under varied Sun directions.

We simulate Sun positions throughout a lunar day using the SPICE toolkit\footnote{\url{https://github.com/AndrewAnnex/SpiceyPy}}.
For each DEM patch, we sample \num{9} distinct solar hours spanning the local daytime period, converting UTC timestamps into Sun position vectors in the Moon-fixed reference frame.
This guarantees that sampled illuminations correspond to physically plausible orbital configurations.
Combined with the learned BRDF, we render appearance images under these solar configurations while preserving realistic shadow boundaries, shading gradients, and photometric consistency.

\textbf{DEM Normalization.}
Although lunar relief spans elevations from \SIrange{-5.6}{7.5}{\kilo\meter} globally~\cite{LRO2008CoordinateSystem}, each DEM crop covers only a small region where height range varies significantly with local morphology.
To ensure consistent input representations, we normalize each DEM crop by subtracting its mean elevation and scaling by the dataset-wide standard deviation. This makes models agnostic to absolute altitude while preserving relative terrain structure.

\textbf{Ground Truth and Annotations.}
Each geometry–appearance sample includes:
i) \textit{DEM patch:} Normalized $128 \times 128$ elevation grid at \SI{5}{\meter\per\px};
ii) \textit{Real appearance image:} Orthorectified LRO observation aligned with the DEM;
iii) \textit{Depth map.} Per-pixel depth expressed in the real image camera view and aligned with the appearance image;
iv) \textit{Surface normal map:} Per-pixel surface normals expressed in the real image camera frame;
v) \textit{Multi-lighting images:} Nine rendered appearances under SPICE-sampled solar positions;
vi) \textit{Illumination metadata:} Sun azimuth, elevation, and UTC timestamp for each image;
vii) \textit{BRDF parameters:} Learned reflectance coefficients for the local terrain;
viii) \textit{Geographic coordinates:} Patch center location in selenographic coordinates;

\textbf{Photometric Perception Tasks.} 
The geometry \vs appearance pairs of {\lunargr} enable training and evaluation of monocular algorithms for depth estimation, crater detection, and semantic segmentation.
The multi-lighting renderings support normal estimation via photometric stereo and serve as principled data augmentation for training illumination-robust models.
%
Additionally, the pixel-wise correspondence between real LRO observations and local DEM geometry provides supervision for reflectance estimation, enabling evaluation and improvement of BRDF prediction methods.

\begin{table*}[!t]
    \centering
    \caption{Stereo 3D reconstruction results on \lunarstereo.
    Best results are highlighted in {\color{mygreen}green} and second-best in {\color{mymyorange}orange}.}
    \label{tab:quantitatif}
    \vspace{-1em}
    
    \setlength{\tabcolsep}{2.3pt}
    \renewcommand{\arraystretch}{1.15}
    
    \scalebox{0.85}{
    \begin{tabular}{ll|ccc|ccc|ccc}
    \toprule
    
    \multirow{2}{*}{\textbf{Dataset}} &
    \multirow{2}{*}{\textbf{Method}} &
    \multicolumn{3}{c|}{\textbf{Nadir}} &
    \multicolumn{3}{c|}{\textbf{Oblique}} &
    \multicolumn{3}{c}{\textbf{Dynamic}} \\
    
    \cmidrule(lr){3-5} \cmidrule(lr){6-8} \cmidrule(lr){9-11}
    & &
    ACC.(m)$\downarrow$ & Compl.(m)$\downarrow$ & Chamfer(m)$\downarrow$ &
    ACC.(m)$\downarrow$ & Compl.(m)$\downarrow$ & Chamfer(m)$\downarrow$ &
    ACC.(m)$\downarrow$ & Compl.(m)$\downarrow$ & Chamfer(m)$\downarrow$ \\
    
    \midrule
    \multirow{4}{*}{\textbf{Tycho (Unseen)}}
    & MASt3R 
    & 60 & 58 & 59
    & 128 & 88 & 108
    & 49 & 45 & 47 \\
    
    & VGGT
    & 39 & 39 & 40
    & 95 & 78 & 87
    & 75 & 51 & 63 \\
    
    & VGGT FT
    & \textbf{\textcolor{mygreen}{35}} & \textbf{\textcolor{mygreen}{35}} & \textbf{\textcolor{mygreen}{35}}
    & \textbf{\textcolor{mymyorange}{76}} & \textbf{\textcolor{mymyorange}{60}} & \textbf{\textcolor{mymyorange}{68}}
    & \textbf{\textcolor{mymyorange}{47}} & \textbf{\textcolor{mymyorange}{52}} & \textbf{\textcolor{mymyorange}{50}} \\
    
    & MASt3R FT
    & \textbf{\textcolor{mymyorange}{37}} & \textbf{\textcolor{mymyorange}{37}} & \textbf{\textcolor{mymyorange}{37}}
    & \textbf{\textcolor{mygreen}{72}} & \textbf{\textcolor{mygreen}{57}} & \textbf{\textcolor{mygreen}{64}}
    & \textbf{\textcolor{mygreen}{34}} & \textbf{\textcolor{mygreen}{32}} & \textbf{\textcolor{mygreen}{33}} \\

    \midrule
    \multirow{4}{*}{\textbf{S. Pole (Seen)}}
    & MASt3R
    & 236 & 235 & 236
    & 385 & 259 & 322
    & 289 & 270 & 279 \\
    
    & VGGT
    & 225 & 228 & 226
    & 222 & 210 & 216
    & 323 & 353 & 300 \\
    
    & VGGT FT
    & \textbf{\textcolor{mygreen}{43}} & \textbf{\textcolor{mygreen}{43}} & \textbf{\textcolor{mygreen}{43}}
    & \textbf{\textcolor{mygreen}{64}} & \textbf{\textcolor{mygreen}{63}} & \textbf{\textcolor{mygreen}{64}}
    & \textbf{\textcolor{mygreen}{69}} & \textbf{\textcolor{mygreen}{69}} & \textbf{\textcolor{mygreen}{70}} \\
    
    & MASt3R FT
    & \textbf{\textcolor{mymyorange}{103}} & \textbf{\textcolor{mymyorange}{97}} & \textbf{\textcolor{mymyorange}{100}}
    & \textbf{\textcolor{mymyorange}{141}} & \textbf{\textcolor{mymyorange}{147}} & \textbf{\textcolor{mymyorange}{144}}
    & \textbf{\textcolor{mymyorange}{109}} & \textbf{\textcolor{mymyorange}{114}} & \textbf{\textcolor{mymyorange}{111}} \\
    
    \bottomrule
    \end{tabular}
}
\end{table*}

\section{{\name} Dataset} \label{sec:dataset}


\textbf{Data Description}. We analyze the characteristics of the sub-datasets in the following.

\begin{figure}
    \centering
    \includegraphics[width=1\linewidth]{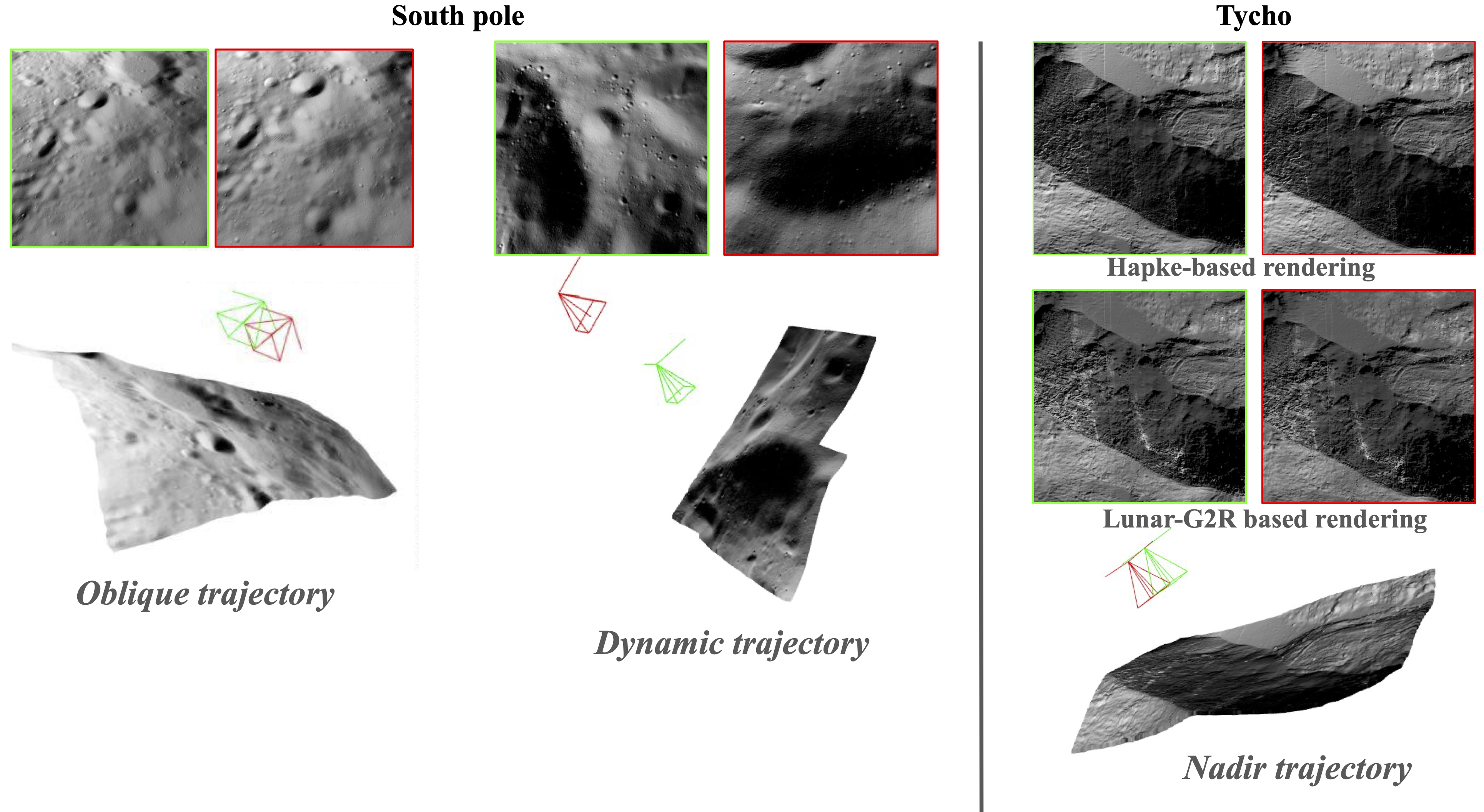}
\caption{Examples of \lunarstereo illustrating the three trajectory types. The first row shows stereo pairs and their associated 3D scene over the lunar South Pole, rendered using the Hapke BRDF with constant parameters. The second row presents two stereo pairs from the Tycho crater: the first rendered using the classical Hapke BRDF, and the second using SVBRDF from~\cite{grethen2026lunar}.}

    \label{fig:LunarGeo_samples}
\end{figure}

\lunarstereo is a synthetic lunar stereo dataset covering two regions: the South Pole and the Tycho crater, with three acquisition trajectories (oblique, dynamic, nadir). 
Stereo pairs for the South Pole use the classical Hapke BRDF. For Tycho, pairs are rendered with both Hapke and a spatially-varying BRDF (SVBRDF), enabling more realistic, crater-specific reflectance.
\cref{fig:LunarGeo_samples} shows example stereo pairs and 3D scenes for the different trajectories and models. 
Each pair includes a dense depth map (EXR) and camera parameters (intrinsic and extrinsic in NPZ), allowing full 3D pose recovery.
\cref{tab:stereolunar_stats} summarizes dataset size and splits: about \num{38}K pairs for the south pole and \num{20}K for Tycho, split \SI{80}{\percent}/\SI{10}{\percent}/\SI{10}{\percent} for train/val/test.

\begin{table}[t]
\centering
\caption{\lunarstereo dataset statistics and train/val/test splits.}
\vspace{-1em}
\label{tab:stereolunar_stats}
\footnotesize
\begin{tabular}{lcccc}
\toprule
\textbf{Region} & \textbf{Total} & \textbf{Train (\SI{80}{\percent})} & \textbf{Val (\SI{10}{\percent})} & \textbf{Test (\SI{10}{\percent})} \\
\midrule
South Pole & 38K & 30K & 4K & 4K \\
Tycho      & 20K & 16K & 2K & 2K \\
\bottomrule
\end{tabular}
\end{table}

\lunarphoto is composed of \num{84}K samples, split into \num{67}K training, \num{8.5}K validation, and \num{8.5}K test samples.
Each sample represents a local lunar surface patch and combines geometric information, real orbital imagery, and physically-based renderings under varying illumination conditions.
For every sample, nine distinct Sun configurations are provided, resulting in a total of \num{750}K rendered images per BRDF model.
\cref{tab:lunarg2r_stats} summarizes the overall dataset statistics and the content provided for each sample.

\begin{figure}
    \centering
    \includegraphics[width=1\linewidth]{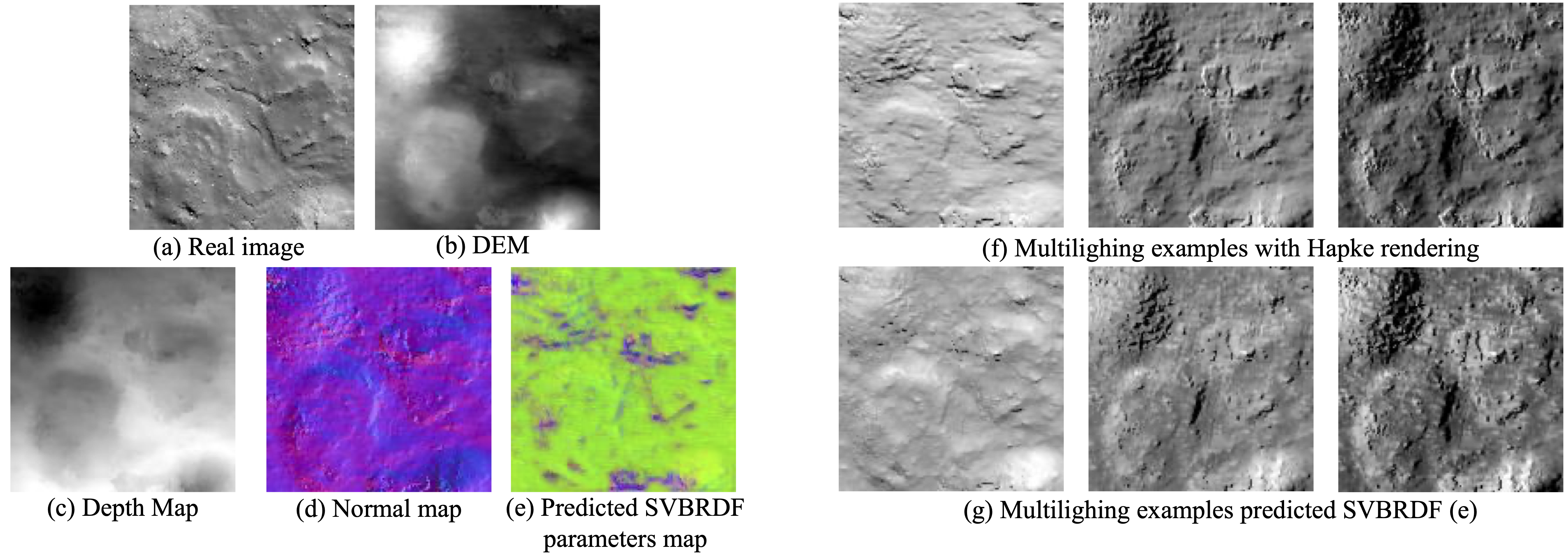}
    \caption{Samples from {\lunargr}.}
    \label{fig:lunarphoto_example}
\end{figure}

\textbf{Qualitative Examples.} 
Examples are shown in \cref{fig:lunarphoto_example}.
Each data sample includes a DEM crop (b) and an orthorectified real lunar image acquired by the LRO NAC (a).
Depth (c) and surface normal maps (d) are provided in the viewpoint of the real image camera, obtained by rendering the image onto the DEM using the corresponding camera parameters.
In addition, per-pixel SVBRDF parameters (e) are supplied together with multi-illumination renderings generated using both the classical Hapke reflectance model (f) and the SVBRDF (g), all stored in TIFF format.
Metadata files include camera parameters, Sun directions, and geolocation information, enabling reproducible rendering and downstream tasks.

\begin{table}[t]
\centering
\caption{\lunarphoto statistics and per-sample content.}
\vspace{-1em}
\label{tab:lunarg2r_stats}
\footnotesize
\begin{tabular}{lp{0.62\columnwidth}}
\toprule
\textbf{Item} & \textbf{Description} \\
\midrule
Total samples & 84K samples \\
Split & 67K train / 8.5K val / 8.5K test \\
Sun positions (multi-lighting) & 9 per sample \\
Rendered images & 750K per BRDF model \\
\midrule
Geometry & \texttt{dem.tif}, \texttt{depth.tif}, \texttt{normal.tif} \\
Real data & \texttt{real\_image.tif} (orthorectified LRO NAC image) \\
Metadata & \texttt{metadata.json} (camera, Sun, geolocation) \\
BRDF & \texttt{brdf\_map.tif} (per-pixel SV-BRDF parameters) \\
Renderings & 
Multi-illumination renderings using Hapke and learned BRDF models (TIFF) \\
\bottomrule
\end{tabular}
\end{table}

\begin{figure}
    \centering
    \includegraphics[width=1\linewidth]{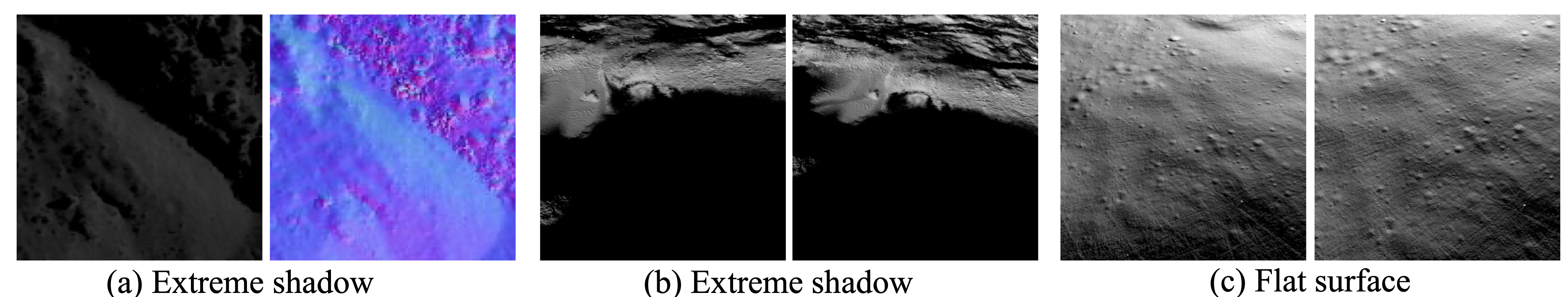}
    \caption{Examples of challenging cases: (a) extreme shadowing observed in the multi-lighting data from \lunargr, (b) a \lunarstereo pair from the Tycho under raking lighting, and (c) a flat stereo pair from the S. Pole captured by \lunarstereo.}
    \label{fig:challenge}
\end{figure}

\textbf{Challenging Scenarios.}
MoonAnything deliberately includes a wide range of challenging visual conditions to promote robustness in downstream vision applications.
In particular, extreme illumination cases with large shadowed regions are common in low-Sun-angle configurations, as illustrated in \ref{fig:challenge} (a) and (b), where strong contrast and partial observability significantly complicate both feature extraction and geometric reasoning.
In addition to illumination variability, the dataset covers diverse terrain types, ranging from highly structured areas such as crater rims, slopes, and ejecta fields to flatter surfaces with limited texture and fewer distinctive features, as shown in (c).
This combination of lighting and terrain diversity is essential to evaluate and train models capable of operating reliably in realistic lunar exploration scenarios.



\section{Stereo 3D Reconstruction as a Showcase} \label{sec:3d_reconstruct}

We evaluate the practical relevance of MoonAnything using stereo 3D reconstruction as a downstream task.
MASt3R~\cite{leroy2024mast3r} and VGGT~\cite{wang2025vggt} are fine-tuned on the South Pole subset of \lunarstereo and evaluated on stereo sequences acquired under three trajectory configurations: nadir, oblique, and dynamic.
We report 3D reconstruction metrics using completeness, accuracy, and Chamfer distance for geometry.
In addition to in-domain evaluation on the South Pole, performance is also assessed on the Tycho crater to evaluate cross-region generalization.
Quantitative results are reported in \cref{tab:quantitatif}.

Fine-tuning on a limited but physically consistent portion of MoonAnything leads to substantial improvements over pretrained models across all trajectory configurations.
On the \textit{seen} South Pole split, both models significantly reduce reconstruction errors, with VGGT~FT (where FT denotes fine-tuning) performing best for nadir views, while MASt3R~FT shows increased robustness under oblique and dynamic trajectories.

Importantly, similar gains are observed on the \textit{unseen} Tycho split, despite its markedly different terrain morphology.
MASt3R~FT achieves the lowest errors for oblique and dynamic trajectories, demonstrating strong generalization to out-of-distribution lunar landscapes.
Overall, these results show that fine-tuning on a restricted subset of MoonAnything is sufficient to learn transferable, geometry-aware representations, validating both the realism of the proposed rendering pipeline and the suitability of MoonAnything for downstream planetary 3D vision tasks.

\section{Conclusion} \label{sec:conclusion}

We presented {\name}, a unified benchmark for lunar surface perception that addresses the lack of comprehensive datasets combining geometric and photometric supervision.
By unifying and extending our prior works~\cite{grethen2025adapting,grethen2026lunar}, {\name} provides stereo imagery of \gmr{large} dataset size with dense depth maps across two lunar regions ({\lunarstereo}) and spatially-varying BRDF parameters with multi-illumination renderings ({\lunargr}). Together, these sub-datasets offer the first benchmark enabling research on 3D reconstruction, reflectance estimation, and illumination-robust perception within a consistent lunar context. The Moon specific setting offered a calibrated (e.g. a unique, known light source, the sun) and large dataset/ground truth that calls for generalization in more complex, general setting.


\begin{acks}
This work was carried out with the support of the \grantsponsor{4000140461}{European Space Agency (ESA)}{} under contract No:~\grantnum{4000140461}{4000140461/23/NL/GLC/my}.
\end{acks}
\bibliographystyle{ACM-Reference-Format}
\balance
\bibliography{strings_abb,reference}

@String{Computer = "{IEEE} Computer" }

@String{Springer = "Springer-Verlag" }

@inproceedings{leroy2024mast3r,
  author       = {Vincent Leroy and
                  Yohann Cabon and
                  J{\'{e}}r{\^{o}}me Revaud},
  title        = {Grounding Image Matching in 3D with MASt3R},
  booktitle    = eccv,
  XXpages        = {71--91},
  year         = {2024},
  XXurl          = {https://doi.org/10.1007/978-3-031-73220-1\_5},
  XXdoi          = {10.1007/978-3-031-73220-1\_5}
}

@inproceedings{wang2025vggt,
  author       = {Jianyuan Wang and
                  Minghao Chen and
                  Nikita Karaev and
                  Andrea Vedaldi and
                  Christian Rupprecht and
                  David Novotn{\'{y}}},
  title        = {{VGGT:} Visual Geometry Grounded Transformer},
  booktitle    = cvpr,
  XXpages        = {5294--5306},
  year         = {2025},
  XXurl          = {https://openaccess.thecvf.com/content/CVPR2025/html/Wang\_VGGT\_Visual\_Geometry\_Grounded\_Transformer\_CVPR\_2025\_paper.html},
  XXdoi          = {10.1109/CVPR52734.2025.00499},
  bibsource    = {dblp computer science bibliography, https://dblp.org}
}

@inproceedings{wang2024dust3r,
  author       = {Shuzhe Wang and
                  Vincent Leroy and
                  Yohann Cabon and
                  Boris Chidlovskii and
                  J{\'{e}}r{\^{o}}me Revaud},
  title        = {DUSt3R: Geometric 3D Vision Made Easy},
  booktitle    = cvpr,
  XXpages        = {20697--20709},
  year         = {2024},
  XXurl          = {https://doi.org/10.1109/CVPR52733.2024.01956},
  XXdoi          = {10.1109/CVPR52733.2024.01956},
  bibsource    = {dblp computer science bibliography, https://dblp.org}
}

@misc{wang2024change,
  author       = {Yanbo Wang and
                  Ting Yuan and
                  Chuankai Liu and
                  Qi Wu and
                  Jiuchao Qian},
  title        = {The Real Chang'e Lunar Landscape Dataset},
  XXpublisher    = {{IEEE} DataPort},
  year         = {2024},
  month        = aug,
  XXhowpublished = {\url{https://doi.org/10.21227/9y8q-dx27}},
  XXnote         = {Accessed on YYYY-MM-DD.},
  XXurl          = {https://doi.org/10.21227/9y8q-dx27},
  doi          = {10.21227/9Y8Q-DX27},
  bibsource    = {dblp computer science bibliography, https://dblp.org}
}

@inproceedings{niles2025commercial,
  title={The Commercial Lunar Payload Services Initiative},
  author={Niles, Paul B and Stephan, Ryan},
  booktitle=aeroconf,
  XXpages={1--7},
  year={2025},
  XXorganization={IEEE}
}

@article{goswami2009chandrayaan,
  title={Chandrayaan-1: India's first planetary science mission to the Moon},
  author={Goswami, JN and Annadurai, M},
  journal={Current science},
  pages={486--491},
  year={2009},
  publisher={JSTOR}
}

@inproceedings{johnson2008overview,
  title={Overview of terrain relative navigation approaches for precise lunar landing},
  author={Johnson, Andrew E and Montgomery, James F},
  booktitle=aeroconf,
  XXpages={1--10},
  year={2008},
  XXorganization={IEEE}
}

@inproceedings{johnson2008analysis,
  title={Analysis of on-board hazard detection and avoidance for safe lunar landing},
  author={Johnson, Andrew E and Huertas, Andres and Werner, Robert A and Montgomery, James F},
  booktitle=aeroconf,
  XXpages={1--9},
  year={2008},
  XXorganization={IEEE}
}

@misc{lebreton2024trainingdatasetsgenerationmachine,
      title={Training Datasets Generation for Machine Learning: Application to Vision Based Navigation}, 
      author={Jérémy Lebreton and Ingo Ahrns and Roland Brochard and Christoph Haskamp and Hans Krüger and Matthieu Le Goff and Nicolas Menga and Nicolas Ollagnier and Ralf Regele and Francesco Capolupo and Massimo Casasco},
      year={2024},
      eprint={2409.11383},
      archivePrefix={arXiv},
      primaryClass={cs.CV},
}

@article{herbort2011introduction,
  title={An introduction to image-based 3D surface reconstruction and a survey of photometric stereo methods},
  author={Herbort, Steffen and W{\"o}hler, Christian},
  journal={3D Research},
  XXvolume={2},
  XXnumber={3},
  XXpages={1--17},
  year={2011},
  XXpublisher={Springer}
}

@article{bickel2021peering,
  title={Peering into lunar permanently shadowed regions with deep learning},
  author={Bickel, Valentin Tertius and Moseley, Ben and Lopez-Francos, I and Shirley, M},
  journal={Nature communications},
  volume={12},
  number={1},
  pages={5607},
  year={2021},
  publisher={Nature Publishing Group UK London}
}

@article{posada2024dense,
  title={Dense Feature Matching for Hazard Detection and Avoidance Using Machine Learning in Complex Unstructured Scenarios},
  author={Posada, Daniel and Henderson, Troy},
  journal={Aerospace},
  volume={11},
  number={5},
  pages={351},
  year={2024},
  publisher={MDPI}
}

@inproceedings{grethen2025adapting,
  title={Adapting Stereo Vision From Objects To 3D Lunar Surface Reconstruction with the StereoLunar Dataset},
  author={Grethen, Cl{\'e}mentine and Gasparini, Simone and Morin, G{\'e}raldine and Lebreton, Jeremy and Marti, Lucas and Sanchez-Gestido, Manuel},
  booktitle=ICCVW,
  XXpages={3751--3760},
  year={2025}
}

@article{grethen2026lunar,
  title={Lunar-G2R: Geometry-to-Reflectance Learning for High-Fidelity Lunar BRDF Estimation},
  author={Grethen, Clementine and Menga, Nicolas and Brochard, Roland and Morin, Geraldine and Gasparini, Simone and Lebreton, Jeremy and Gestido, Manuel Sanchez},
  journal={arXiv:2601.10449},
  year={2026}
}

@article{vondrak2010lunar,
  title={Lunar Reconnaissance Orbiter (LRO): Observations for lunar exploration and science},
  author={Vondrak, Richard and Keller, John and Chin, Gordon and Garvin, James},
  journal={Space science reviews},
  volume={150},
  number={1},
  pages={7--22},
  year={2010},
  publisher={Springer}
}

@article{Barker2016,
  title = {A new lunar digital elevation model from the Lunar Orbiter Laser Altimeter and {SELENE} Terrain Camera},
  volume = {273},
  ISSN = {0019-1035},
  url = {http://dx.doi.org/10.1016/j.icarus.2015.07.039},
  DOI = {10.1016/j.icarus.2015.07.039},
  journal = {Icarus},
  publisher = {Elsevier BV},
  author = {Barker,  M.K. and Mazarico,  E. and Neumann,  G.A. and Zuber,  M.T. and Haruyama,  J. and Smith,  D.E.},
  year = {2016},
  month = jul,
  pages = {346–355}
}

@article{liu2024lusnar,
  author       = {Jiayi Liu and
                  Qianyu Zhang and
                  Xue Wan and
                  Shengyang Zhang and
                  Yaolin Tian and
                  Haodong Han and
                  Yutao Zhao and
                  Baichuan Liu and
                  Zeyuan Zhao and
                  Xubo Luo},
  title        = {LuSNAR:A Lunar Segmentation, Navigation and Reconstruction Dataset
                  based on Muti-sensor for Autonomous Exploration},
  journal      = {arXiv 2407.06512},
  XXvolume       = {abs/2407.06512},
  year         = {2024},
  XXurl          = {https://doi.org/10.48550/arXiv.2407.06512},
  Xdoi          = {10.48550/ARXIV.2407.06512},
  eprinttype    = {arXiv},
  XXeprint       = {2407.06512},
  bibsource    = {dblp computer science bibliography, https://dblp.org}
}

@article{wong2017polar,
  title={Polar optical lunar analog reconstruction (polar) stereo dataset},
  author={Wong, Uland and Nefian, Ara and Edwards, Larry and Buoyssounouse, Xavier and Furlong, P Michael and Deans, Matt and Fong, Terry},
  journal={NASA Ames Research Center},
  year={2017}
}

@article{marcus2024slt,
  author       = {Marcus M{\"{a}}rtens and
                  Kevin Farries and
                  John Culton and
                  Tat{-}Jun Chin},
  title        = {Synthetic Lunar Terrain: {A} Multimodal Open Dataset for Training
                  and Evaluating Neuromorphic Vision Algorithms},
  journal      = {arXiv 2408.16971},
  XXvolume       = {abs/2408.16971},
  year         = {2024},
  XXurl          = {https://doi.org/10.48550/arXiv.2408.16971},
  XXdoi          = {10.48550/ARXIV.2408.16971},
  XXeprinttype    = {arXiv},
  XXeprint       = {2408.16971}
}

@misc{ISRO_Chandrayaan2,
  author       = {{Indian Space Research Organisation (ISRO)}},
  title        = {Chandrayaan-II},
  howpublished = {\url{https://pradan.issdc.gov.in/ch2/}},
  xxnote         = {Accessed: 3 January 2026}
}

@misc{pessia2019artificial,
  author       = {Pessia, Romain and Ishigami, Genya and Jodelet, Quentin},
  title        = {Artificial Lunar Landscape Dataset},
  year         = {2019},
  howpublished = {Dataset}
}

@inproceedings{parkes2004pangu,
  title={Planet surface simulation with pangu},
  author={Parkes, SM and Martin, Iain and Dunstan, Martin and Matthews, D},
  booktitle=spaceops,
  XXpages={389},
  year={2004}
}

@article{lebreton2024high,
  title={High performance Lunar landing simulations},
  author={Lebreton, J{\'e}r{\'e}my and Brochard, Roland and Ollagnier, Nicolas and Baudry, Matthieu and Salah, Adrien Hadj and Jonniaux, Gr{\'e}gory and Kanani, Keyvan and Goff, Matthieu Le and Masson, Aurore},
  journal={arXiv:2409.11450},
  year={2024}
}

@misc{robinson2011lro,
  author       = {Robinson, Mark},
  title        = {{LRO Moon LROC 5 RDR v1.0}},
  year         = {2011},
  note         = {NASA Lunar Reconnaissance Orbiter Camera (LROC) data product}
}

@book{hapke1993theory,
  author = {Hapke, Bruce},
  title = {{Theory of reflectance and emittance spectroscopy}},
  publisher = cup,
  year = {1993},
  XXaddress = {Cambridge, UK}
}

@article{YU2014,
  title = {Guidance navigation and control for {Chang'E-3} powered descent},
  volume = {44},
  ISSN = {1674-7259},
  XXurl = {http://dx.doi.org/10.1360/092014-43},
  XXDOI = {10.1360/092014-43},
  XXnumber = {4},
  journal = sst,
  XXpublisher = {Science China Press.,  Co. Ltd.},
  author = {Yu,  Ping and Zhao, Yu and Li, Ji and Zhang, XiaoWen and Wang,  DaYi and Huang, XiangYu and Liang, Jun and Guan,  YiFeng and Zhang, HongHua and Yang, Wei},
  year = {2014},
  XXmonth = apr,
  XXpages = {377–384}
}

@article{Wong2006,
  title = {Guidance and Control Design for Hazard Avoidance and Safe Landing on Mars},
  XXvolume = {43},
  ISSN = {1533-6794},
  XXurl = {http://dx.doi.org/10.2514/1.19220},
  XXDOI = {10.2514/1.19220},
  XXnumber = {2},
  journal = jspacrock,
  XXpublisher = {American Institute of Aeronautics and Astronautics (AIAA)},
  author = {Wong,  Edward C. and Singh,  Gurkirpal and Masciarelli,  James P.},
  year = {2006},
  XXmonth = mar,
  XXpages = {378–384}
}

@inproceedings{Getchius2024,
  title = {{Hazard Detection and Avoidance for the Nova-C Lander}},
  ISBN = {9783031519284},
  ISSN = {2731-0884},
  XXDOI = {10.1007/978-3-031-51928-4_53},
  booktitle = AASGNC,
  author = {Getchius,  Joel and Renshaw,  Devin and Posada,  Daniel and Henderson,  Troy and Hong,  Lillian and Ge,  Shen and Molina,  Giovanni},
  year = {2024},
  XXpages = {921–943}
}

@techreport{LRO2008CoordinateSystem,
  author      = {{LRO Project}},
  title       = {A Standardized Lunar Coordinate System for the Lunar Reconnaissance Orbiter},
  institution = {NASA},
  year        = {2008},
  XXtype        = {Technical Report},
  XXnote        = {LRO Project White Paper, Version 4}
}

@article{sato2014hapke,
  title={Resolved Hapke parameter maps of the Moon},
  author={Sato, Hiroyuki and Robinson, Mark S. and Hapke, Bruce and Denevi, Brett W. and Boyd, Angela K.},
  journal=jgr,
  volume={119},
  number={8},
  XXpages={1775--1805},
  year={2014},
  XXpublisher={Wiley},
}

@STRING{cvpr = "{CVPR}"}

@STRING{eccv = "{ECCV}"}

@STRING{iccvw = "ICCVW"}

@string{AASGNC = {AASG\&C}}

@string{cup = {CUP}}

@string{aeroconf = {AeroConf}}

@string{spaceops ={SpaceOps}}

@string{jgr = {J. Geophys. Res.}}

@string{jspacrock = {J. Spacecr. Rockets}}

@string{sst = {Sci. Sin. Technol.}}

\end{document}